\documentclass[conference]{IEEEtran}
\usepackage{cite}
\usepackage{amsmath,amssymb,amsfonts}
\usepackage{algorithmic}
\usepackage{graphicx}
\usepackage{textcomp}
\usepackage{xcolor}
\usepackage{tabularx,booktabs}
\usepackage{url}
\newcolumntype{C}{>{\centering\arraybackslash}X} 
\def\BibTeX{{\rm B\kern-.05em{\sc i\kern-.025em b}\kern-.08em
    T\kern-.1667em\lower.7ex\hbox{E}\kern-.125emX}}
\begin{document}

\title{Maintainability Challenges in ML: A Systematic Literature Review}

\author{\IEEEauthorblockN{Karthik Shivashankar}
\IEEEauthorblockA{\textit{Department of Informatics},
\textit{University of Oslo, Norway}\\
karths@ifi.uio.no}
\and
\IEEEauthorblockN{Antonio Martini}
\IEEEauthorblockA{\textit{Department of Informatics},
\textit{University of Oslo, Norway}\\
antonima@ifi.uio.no}
}
\maketitle
\begin{abstract}
\textit{Background}: As Machine Learning (ML) advances rapidly in many fields, it is being adopted by academics and businesses alike. However, ML has a number of different challenges in terms of maintenance not found in traditional software projects. Identifying what causes these maintainability challenges can help mitigate them early and continue delivering value in the long run without degrading ML performance. \textit{Aim}: This study aims to identify and synthesise the maintainability challenges in different stages of the ML workflow and understand how these stages are interdependent and impact each other's maintainability. \textit{Method}: Using a systematic literature review, we screened more than 13000 papers, then selected and qualitatively analysed 56 of them.\textit{ Results}:  (i) a catalogue of maintainability challenges in different stages of Data Engineering, Model Engineering workflows and the current challenges when building ML systems are discussed; (ii) a map of 13 maintainability challenges to different interdependent stages of ML that impact the overall workflow; (iii) Provided insights to developers of ML tools and researchers. \textit{Conclusions}: In this study, practitioners and organisations will learn about maintainability challenges and their impact at different stages of ML workflow. This will enable them to avoid pitfalls and help to build a maintainable ML system. The implications and challenges will also serve as a basis for future research to strengthen our understanding of the ML system's maintainability.
\end{abstract}
\begin{IEEEkeywords}
Machine Learning, Deep Learning, Artificial Intelligence, Maintainability, Systematic Literature Review
\end{IEEEkeywords}
\section{Introduction}
Modern Software applications rely heavily on Machine learning (ML) systems and are used in various tasks to provide meaningful insights learned from growing and evolving data. Many companies have adapted ML in their service offering and also delivered value internally. The increasing adoption of ML has introduced new challenges associated with data management and processing, model training and deployments, data and model quality assurance, and other development practices \cite{c2}. \\ML systems are data-driven and data-dependent, thereby creating entanglement between data features and model performance, making them susceptible to model staleness and training-serving skew, which may degrade the performance of the ML system without proper mitigation strategies in place\cite{c10}. Furthermore, this data dependency behaviour costs more than code dependency, making them particularly vulnerable to Technical debt (TD) and incurring massive ongoing maintenance costs compared to traditional software projects\cite{c17}. \\ Hence, it is imperative that organisations and practitioners understand how to develop maintainable ML systems and how they can continue to deliver value in the long run. The first step towards overcoming this problem is understanding maintainability challenges when developing ML systems and how different stages in the workflow affect their maintenance. Unfortunately, no systematic literature review has investigated maintainability challenges in ML systems despite their importance. To fill the gap in the current literature, we conducted a Systematic Literature Review (SLR) on Maintainability challenges in ML systems. We have the following Research Questions (RQ), which guided our SLR study. \textbf{(RQ1)} What are the  Data Engineering Maintainability challenges? \textbf{(RQ2)} What are the  Model Engineering Maintainability challenges? \textbf{(RQ3)} What are the  current maintainability challenges when Building an ML systems?
We have selected 56 primary studies for inclusion in this SLR. The main contributions of our SLR study are as follows.
\textbf{Contribution 1:} ML maintainability challenges which are identified and discussed using our \textbf{RQ's} will help the community to make prudent choices when developing or maintaining ML system or application. 
\textbf{Contribution 2:} Mapping of the 13  synthesised interdependent stages and its impact on maintainability will guide the practitioner to be wary of the dependencies and cost involved during maintenance when dealing with each stages.
 \textbf{Contribution 3:} We have also  distilled implications for Developers of ML tools and synthesised opportunity for further research.
\section{Background}
\subsection{Maintainability of Software Systems}
Software maintainability means "the ease with which a software system or component can be modified to correct faults, improve performance or other attributes and adapt to a changing environment"\cite{c5}. Software systems are frequently changed to meet changing customer requirements that may arise from various factors, including changes in technology or enhancing existing features \cite{c16}. Compared to development, software maintenance consumes more resources, effort, and time. It is estimated that Software developers spend about 70\% of their time on maintenance \cite{c12}. There is also a  high degree of complexity in today's software, and the size of the software has grown considerably, making maintenance increasingly difficult \cite{c13}. An organisation's and a product's success relies directly on its software maintainability \cite{c14}. Therefore, producing "software that is easy to maintain" could save a lot of time and money and deliver long term value.
\subsection{Data and Model Engineering}
In ML, data is the first-class citizen and it is well known that the majority of the time spent on ML development is spent on processing data \cite{b12}. ML algorithms cannot perform well without handling dirty data since data quality profoundly impacts model accuracy. ML workflows usually begin with acquiring and preparing the data for training. Creating high-quality training data is typically a tedious, repetitious process\cite{b14}. Data engineering pipelines typically involve a sequence of operations on a set of data from various sources. These operations aim to create training and testing datasets for the ML algorithms. Generally, data engineering is divided into many stages: Data acquisition and exploration, Data processing, Data validation and management \cite{c15}. \\
Model training is the process of feeding an algorithm with data to learn patterns instead of having to manually discover and encode those patterns \cite{c1}. A model engineering pipeline consists of several operations that result in a final model usually used by ML engineers and data science teams. These operations include Model Training, Hyper-Parameter Optimization (HPO), Model Governance, Model Monitoring, Model Testing, Model drift, and Model Deployment\cite{c15}.
\subsection{Related Works}
Numerous studies have identified the different types of TDs, mainly an extension of Sculley et. al.\cite{c17}, and anti-patterns that emerge in the development of ML systems, and how they affect Model performance \cite{c10}. Furthermore, we found studies discussing challenges related to applying existing SE techniques to ML development, using a case study approach, and an empirical study discussing the use of mature engineering techniques to increase reliance on ML components\cite{c3,c7}. We also found a paper that examines the challenges associated with ML deployment and provides a framework for accelerating ML development\cite{c6} and architectural challenges for ML systems\cite{c2}. Another study identifies and categorises data management challenges faced by practitioners at different stages of ML workflow\cite{c9}. Nevertheless, all these papers do not address the different maintainability challenges that arise at various stages of ML workflow, for example, how the data-dependent and stochastic nature of ML affects the maintainability of model testing, data validation, and detecting model drifts. We seek to fill this gap in research by synthesising the maintainability challenges at each stage in the ML workflow and how they are interdependent and impact each other using an SLR approach.
\section{Methodology}
The following section details our rigorous strategy, closely adhering to the guidelines suggested for conducting SLRs by Kitchenham et al.\cite{c8}. We also provide a\textbf{ replication package}, available at this  \url{https://doi.org/10.5281/zenodo.6400559}{ Zenodo link}.
\subsection{Research method}
Fig. 1. shows an overview of how this SLR study was conducted. The selected databases were screened using well-defined search terms and queries to obtain the desired papers.
The resultant papers are then evaluated iteratively based on inclusion and exclusion criteria. The categorization of unclear and conflicting papers was reviewed by the second author.
\begin{figure}[htbp]
\centerline{\includegraphics[scale=0.75]{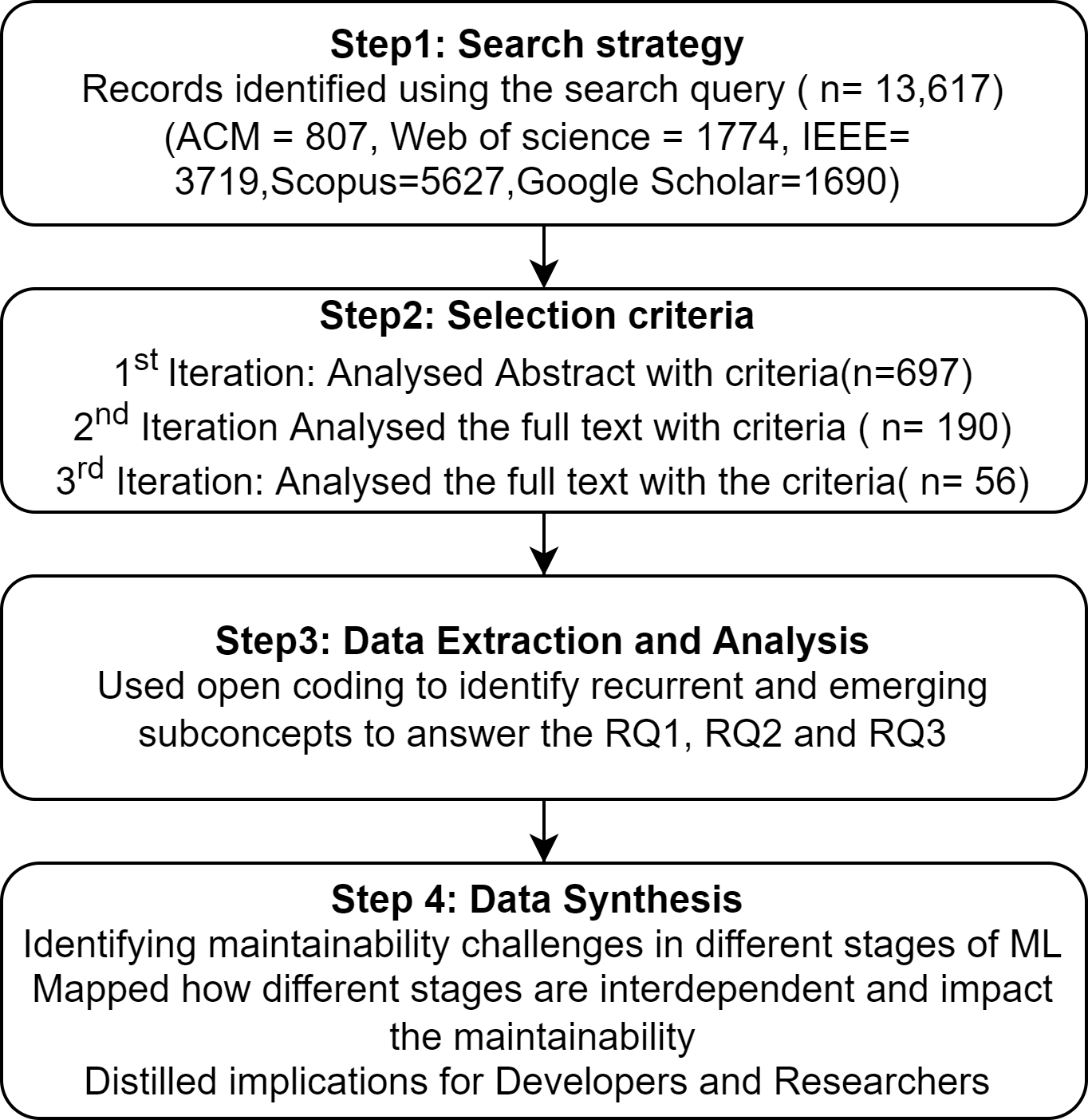}}
\caption{Systematic Literature Review Process}
\label{fig}
\end{figure}
\subsection{Step 1: Search strategy:}
Databases included in this Systematic Literature Review (SLR) are IEEE Xplore, ACM Digital Library, Web of Science, Google Scholar and Scopus. Keywords contained in the title, abstract and index terms of the literature are identified by the following search query for most of the databases which facilitated these features see the replication package Zenodo link for the exact search query used for different databases.
\begin{itemize}
\item  \textbf{TITLE-ABS-KEY} ( machine   \textbf{AND}  learning   \textbf{AND} software   \textbf{AND}  ( ml   \textbf{OR}  ai   \textbf{OR}  dl   \textbf{OR}  neural   \textbf{OR}  intelligence   \textbf{OR}  learning )   \textbf{AND}  ( adapt*   \textbf{OR}  maintain*   \textbf{OR}  scal*   \textbf{OR}  exten*   \textbf{OR}   evol*   \textbf{OR}  flex* )   \textbf{AND}  ( system   \textbf{OR}  architect*   \textbf{OR}  design   \textbf{OR}  build   \textbf{OR}  application   \textbf{OR}  engineering   \textbf{OR}  test )   \textbf{AND}  ( data   \textbf{OR}  algorithm   \textbf{OR}  debt   \textbf{OR}  pattern   \textbf{OR}  code ) )
\end{itemize}
The query contains all primary keywords from research questions with alternative spellings, synonyms, and keywords obtained from pilot searches and related papers.\\
The search was conducted for articles published between January 1, 2014, and January 15, 2022. We chose to start from 2014 because the adoption and development of many ML projects and libraries started around that year\cite{c4}. We limited our search to the first 1000 papers from Google Scholar and the first 2000 papers from Scopus in order by relevance, due to a limitation in their respective systems. 
\subsection{Step 2: Selection criteria:}
We performed three iterations using the inclusion and exclusion criteria as shown in Fig. 1. In the first iteration, we analysed the paper's title and abstract, which resulted in 697 papers for the next iterations;  at the end, the second iteration a total of 190 papers were selected and for the final third iteration a total of 56 papers were selected for inclusion in this SLR study.\\
\textbf{Inclusion criteria}
\begin{itemize}
\item Paper that answers at least one Research questions (RQ).
\item Paper focuses on the maintainability aspect of the ML system.
\end{itemize}
\textbf{Exclusion criteria }
\begin{itemize}
\item Research papers that are not written in English language.
\item Publication for which the full text is not available.
\item Grey literature.
\item Duplicate papers and shorter version of already included publications.
\end{itemize}
\subsection{Step 3: Data Extraction and Analysis}
Our next step was to analyse the selected literature and extract data related to our research question. A paper selected from the literature was studied in-depth and assigned to one or more of the three RQ. Using the open coding technique, recurrent concepts were systematically identified \cite{c11}. Additional axial coding was required to reduce the growing complexity of some emerging concepts (e.g. different stages in data preprocessing like cleaning, splitting and other transformation steps ) \cite{c11}. The authors frequently discussed emerging results to maintain code consistency and high abstraction levels. The emerging subtopics associated with each RQ are discussed in detail in Section IV.
\subsection{Step 4: Data Synthesis}
Our data analysis reports the maintainability challenges associated with different stages of the ML development process in Section IV. In our synthesis step, we identified how these different stages in ML development are interdependent and influence the maintainability of each other (Section V, Table 1). We then model such relationships using a  mapping diagram, as shown in Fig. 2. Finally, in Section VI, we distilled new insights and a roadmap for ML tools developers and researchers as a final synthesis step. In order to ensure transparency and reproducibility, we also made all study artefacts publicly available at \url{https://doi.org/10.5281/zenodo.6400559} .
\subsection{Threats to Validity}
The internal validity is affected by potentially hidden bias, which can affect the consistency and accuracy of the results. SLR synthesis may be biased since it relies on subjective interpretation, especially in mapping Section V, where the author adds a layer of interpretation over the identified maintainability challenges. Although we followed our SLR protocol closely and resolved any conflict between the two authors, other researchers may have obtained slightly different results. External validity concerns the generalisability of the research study. Although we reviewed all the existing research studies, our findings are limited to the results from the final 56 primary studies. Consequently, the maintainability challenges from this SLR may not apply to all scenarios.
\section{Results}
Our analysis summarises the maintainability challenges associated with different stages of  ML workflow. In subsection A, 18 papers were reviewed for the RQ1; in subsection B, 32 papers were reviewed for the RQ2, and in subsection C, 21 papers were reviewed for the RQ3.
\subsection{Data Engineering Maintainability Challenges}
\indent \textbf{Dataset creation} is a manual, slow and error-prone process with inherent bias associated with the data or its collection strategy, which affects the overall performance and quality of the model \cite{b1,b8,b14}. In addition to that, lack of ownership, documentation and transparency in the creation process also undermines its quality\cite{b15}. Usually, datasets are susceptible to missing data, outliers, adversarial and poisoned data \cite{b16} and need to be handled using appropriate data processing strategies with ongoing maintenance as the model is continuously being updated with new data to avoid degradation in ML performance.\\
\indent \textbf{Data preprocessing} pipeline handles data errors like missing data, outliers, lack of metadata, adversarial data and other quality attributes like bias and unfairness associated with the dataset to prepare the data for training\cite{b1,b2}. Furthermore, the model's performance and data features are entangled, so even minor changes in the data feature, like handling missing data or the choice of data splitting strategy, will affect the model's accuracy\cite{b5,b6,b7,b9,b10}. ML workflow being an iterative process, poor model performance or accuracy may often necessitate reevaluating the choice of the data preprocessing steps and handling changes at different stages in the data pipeline, usually in a trial and error manner, which is a waste of resource and time.\\
\indent \textbf{Data management} process includes data acquisition and integration from multiple sources, managing and facilitating manipulation of different modalities of data, modifying annotation or labelling, object serialization and also storing multiple formats of the data. The large scale nature of the data, particularly in Deep Learning (DL), makes this process quite complex and challenging when dealing with an actively evolving dataset\cite{b3}.In addition to that, the highly experimental nature of the ML project also demands provenance tracking, indexing, tracking data transformation steps, and storing intermediary results to ensure reproducibility and reuse of the processed data in the ML workflow. Most of these capabilities require significant maintenance effort and complex engineering and DevOps solutions\cite{b4,b13}.\\
\indent \textbf{Data validation} challenges are profound when data may change as it evolves and error due to possible bugs in the data source \cite{b2} consequently making it complex to monitor and validate what is happening in the data. Most ML  models are complex black boxes, so it becomes unclear whether the learned model still effectively solves the intended use case\cite{b11}. The data validation pipeline will continuously check and monitor for data errors. However, it is pretty challenging to set up and demand substantial engineering resources for its development and maintenance; most engineering teams choose to ignore it in their workflow if it is not a requirement\cite{b12}.
\subsection{Model Engineering Maintainability Challenges}
\indent \textbf{HPO}: Finding an optimal Hyper-parameter is a prolonged process; without expert knowledge, it is often done on a trial and error basis. Because models performance, efficiency and rate of convergence of models are all dependent on HPO\cite{b22,b23}. Wrong choices in these parameters often directly influence the learnability and rate of coverage in the Model training stage, which may often lead to retraining the model with different parameters. Many techniques are available for automated HPO like Bayesian optimisation, Meta-learning and Neural Architecture Search (NAS). Automating HPO requires setting up and maintaining an orchestration pipeline to run the optimisation and to keep track of these parameters and results for workflow reproducibility\cite{b20,b21}.\\
\indent \textbf{Model training}: Maintainability challenges associated with Model training are setting up the infrastructure to automate the training pipeline and monitor the model performance for every iteration. The training utilises extensive computational resources and is very time consuming and costly, especially for Deep Neural Networks\cite{b32}. It is often required to retrain with new data constantly to keep the model updated. Even the choice of model training techniques like incremental training and federated learning will add to the complexity of managing, integrating and deploying the training pipeline to other systems and applications\cite{b18,b19}.\\
\indent \textbf{Model testing} challenges are mainly due to the stochastic nature of ML, rapidly changing input and expected output parts of test cases, oracle issues and emergent functional behaviour, which creates a moving target. Therefore, they are fundamentally different from traditional software projects. As a result, posing new challenges for authoring and maintaining unit tests and regression tests \cite{b33,b34,b35,b36,b37}. Finally, fault testing is also difficult to manage in ML when the learning is based on training data which makes it hard to interpret results from a complex model \cite{b55,b56}. There are still many challenges and open problems related to ML testing and its maintainability.\\
\indent \textbf{Model deployment} challenges arise when transitioning from the test or prototype stage to the production stage, where the model may be deployed and integrated with other models or applications in a different environment set up. These challenges include maintaining glue code, set up monitoring, logging and handling feedback loops \cite{b9,b10,b27}. In addition to that, based on the requirement, it may be challenging to deploy models when the memory and power are a constraint on different platforms ( Mobile or edge device) . This may often result in many maintainability challenges, and deployment issues like model conversion, platform support, vendor-specific optimisation libraries and packages and other interoperability issues\cite{b26,b28,b29}.\\
\indent \textbf{Model drift} are caused by many factors such as data seasonality or changing drift types, evolving data source, and fluctuation in data collection\cite{b5,b17}. All these factors may lead to model staleness and degradation in performances. Most of the methods for detecting drift are expensive to implement because they require knowledge of drift detection algorithms, engineering the solution into existing pipelines, and ongoing maintenance to detect new drifts because it is not possible for the algorithm to detect all drift \cite{b17,b19}.\\
\indent \textbf{Model monitoring } maintainability challenges are caused by evolving input data, fine-grained nature of the quality metrics, prediction bias, and understanding what are the critical metrics of data and model to monitor and how to alert on them \cite{b19,b24}. Furthermore, ML applications in production can also influence their behaviour over time and may lead to undesired feedback loops. Engineers have to build and maintain custom solutions in order to monitor the ML application effectively, with an orchestration pipeline, centralised dashboards for performance monitoring and governance, detecting feedback loops and continuously monitoring the retrained model \cite{b26,b27}. Systems logs are another means to monitor in an ML system, where log entries are typically created in an ad hoc, unstructured and uncoordinated fashion, thereby limiting their usefulness\cite{b25}.\\
 \indent \textbf{Model governance}: It is common for high-risk ML applications to involve cross-disciplinary efforts to define quality metrics and requirements for monitoring the production environment, as well to access its quality in a real-life setting \cite{b19}. However, the stochastic nature of  ML systems make the process painstaking hard to document and manage the risk of the model and to ensure compliance with all regulations and minimum standards\cite{b26}. There is also little guidance for sharing and version controlling ML models and their artefacts such as weights, hyper-parameters, and training and testing sets. Researchers often share ML models through customised websites or GitHub because there are no standard methods. Without publishing these artefacts, it is almost impossible to verify or build upon published results \cite{b30,b31} which impacts the reproducibility and verification of models.
\subsection{Current Maintainability Challenges in Building a ML system}
\indent \textbf{Architecture of ML system}: Current ML solutions do not meet the needs of practitioners \cite{b24,b41,b45}, more framework agnostic, easy to use tooling is needed to ensure the model can be maintained and updated from prototype to production stage\cite{b38}. In addition, there is a disconnect between theory and practice when it comes to data processing, model building,  quality assurance and how to maintain of ML systems \cite{b39,b40}. Most platforms tend to support only one model framework, leading to a tight coupling between the modelling and infrastructure layers. Consequently, practitioners are limited in their ability to develop models and prevented from exploring and using cutting-edge algorithms\cite{b41}. In general, cloud providers do not think about providing a standard programming model that makes ML practitioners' lives easier; they typically use either a black box or a complex runtime environment to approach ML, which offers simplicity at the cost of flexibility \cite{b43}.\\
\indent \textbf{Quality of ML system}: ML has unique quality attributes concerns during development, such as data-dependent behaviour, detecting and responding to drift over time, handling bias, and timely capture of ground truth for retraining of a model to deliver a quality ML system \cite{b42,b50,b52}. Additional, quality concerns include a lack of specifications, defined standards, and documentation in ML workflow and an inability to communicate about model quality due to a lack of a common language \cite{b51,b52}. Maintaining the quality is challenged when ML systems are built to adapt to new situations and contexts, which raises uncertainties regarding the runtime product quality and dependability in an evolving ML system\cite{b35,b53}.\\
\indent \textbf{AutoML}: Non-experts have no idea which of the many ML algorithms to use in order to achieve good performance. AutoML alleviates this challenge by automating model selection and hyper-tuning\cite{b44,b46}. In practice, however, most existing AutoML systems ignore the important stage of data processing\cite{b49}. Therefore, it is hard to minimise expert intervention easily with current computing technologies because developers need to understand how to perform feature engineering, data processing and evaluate bias, interpretation of the model performance and ongoing maintenance cost when dealing with TD like hidden-feedback loops\cite{b47}.\\
\indent \textbf{MLOps}: In practice, engineers spend significant effort developing ad hoc programs for new problems by writing glue code to connect components from different software libraries, processing different forms of raw input, and interfacing with external systems. All these steps are tedious and error-prone and lead to the emergence of brittle pipeline jungles\cite{b27,b45,b48} which are hard to maintain in an MLOps setup. Additionally, using MLOps in a multi-organisation context creates the usual integration problems that emerge in APIs, data formats, privacy, and security, especially from the perspective of governance, auditing, and regulations\cite{b54} which need to be maintained with custom solutions on an ongoing basis.
\section{Interdependence of Maintainability}
The synthesis from our data analysis shows how different stages in the ML workflow affect the maintainability of others (arrows visible in Fig. 2 and Table 1). A single arrow starting from one stage and pointing to another shows how the former stage impacts the maintainability of the latter. Two-way arrows indicate that the two ML stages impact each other. The \textbf{number} associated with each arrow in Fig 2. is referenced by \textbf{row no} in Table 1, where we explain the details and the sources from the SLR. This model will guide practitioners when evaluating the dependencies and maintenance costs for each stage of an ML system life-cycle.\\
Our analysis also reveals an anti-pattern, which we call \textbf{Repetitive Maintenance} (blue oval with arrows in Fig. 2): Poor model accuracy or performance and other quality issues observed in Model testing\cite{b33,b34,b35,b36,b37}, governance\cite{b19,b26} and monitoring\cite{b19,b24} stages may necessitate multiple costly  modification and reevaluations of the steps in dataset creation\cite{b8,b14}, data preprocessing\cite{b5,b6,b7,b9,b10}, data management\cite{b3}, data validation\cite{b2,b11}, HPO\cite{b22,b23}, and model training\cite{b18,b19}.
\begin{figure}[htbp]
\centerline{\includegraphics[scale=0.75]{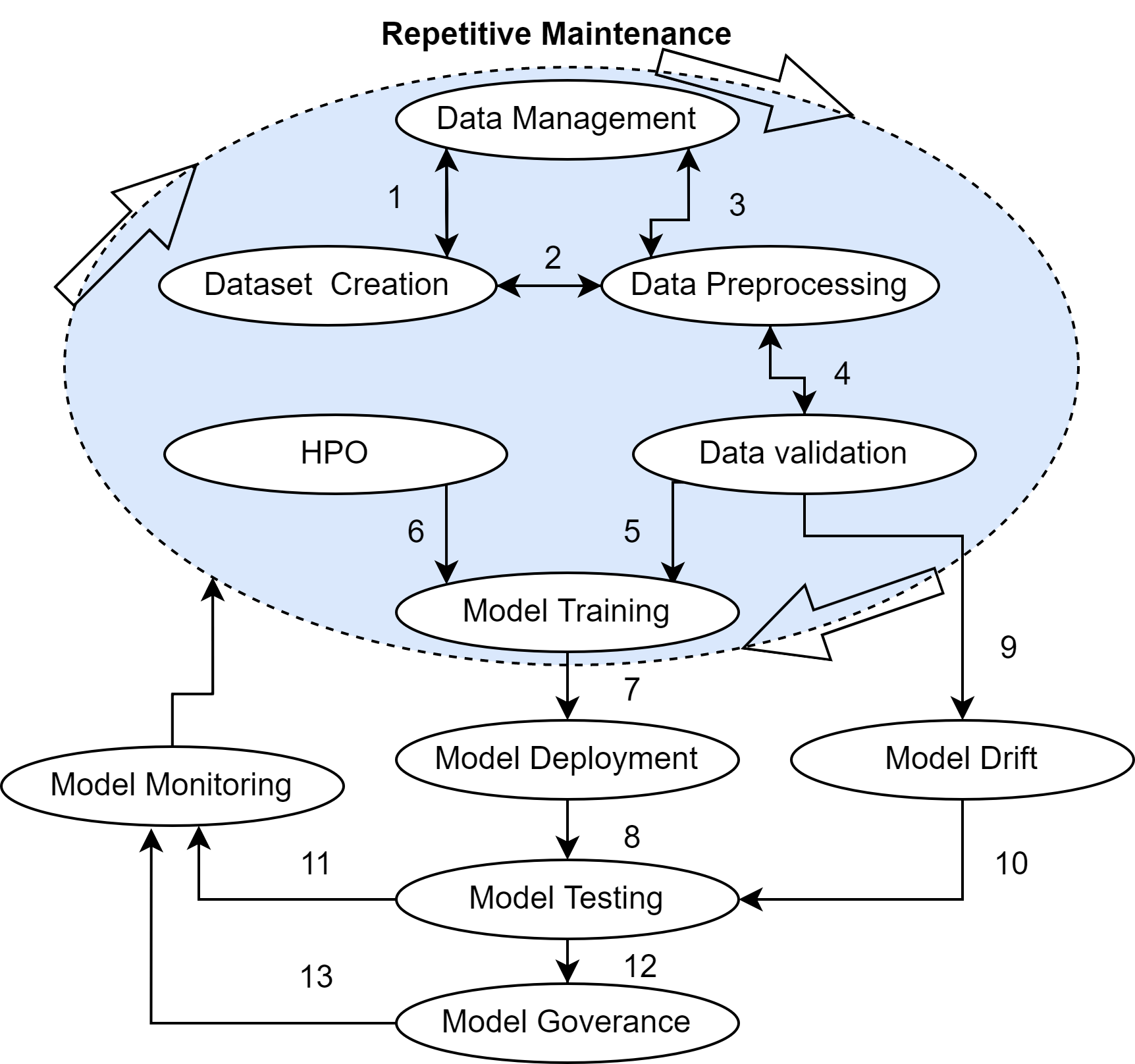}}
\caption{Mapping interdependence of Maintainability challenges in different stages of the ML life-cycle, refer Table 1 for relations}
\label{fig}
\end{figure}
\begin{table*}
 \caption{Maintainability challenges and their interdependence }
\label{my-label}
\begin{tabularx}{\textwidth}{p{.2cm} | p{3cm} | p{13.8cm} }
\toprule
\textbf{No} &  \textbf{Interdependent} \textbf{Stages}  & \textbf{Impact on Maintainability and its challenges} 
\\
\midrule
1 & DatasetCreation  $\longleftrightarrow$
	 Data Management & \scriptsize When dealing with evolving datasets, these stages are interconnected because there is constant back and forth maintenance between them, with data acquisition from different sources, facilitation and manipulation of various types of data, serialization, and storage of multi-format data for reuse further down the pipeline\cite{b3}.\\
\midrule
2 &  DatasetCreation $\longleftrightarrow$	 Data Preprocessing & \scriptsize Dataset can have many errors and quality issues like missing data, outliers, unfairness and could also be adversarial and poisoned \cite{b1,b14,b16,b18}. Consequently, the dataset goes to a preprocessing pipeline where it is cleaned and transformed \cite{b1,b2,b5,b6,b9}  before being used in the training pipeline. Model performance and dataset features are deeply entangled, so even minor changes in the dataset will have an direct consequence to model performance \cite{b5,b6,b7,b9,b10}. Consequently, resulting in the Repetitive Maintenance anti-pattern.\\
\midrule
3 & Data Preprocessing $\longleftrightarrow$	 Data Management & \scriptsize The highly experimental nature of the data preprocessing steps demands provenance tracking, querying and storing transformation steps and intermediary results to ensure reproducibility and reuse. However due to the data-dependent behaviour and entanglement between the data features and model performance makes it challenging to maintain and may lead to correction cascades \cite{b4,b13}.\\
\midrule
4 & Data Preprocessing $\longleftrightarrow$ Data Validation & \scriptsize	 Typical data processing challenges are 1) data may be dirty, 2) data may change as it evolves, 3) error due to possible bugs in the data source \cite{b2}; because of this stochastic nature removing all data errors in preprocessing stage is challenging. Therefore needs a validation pipeline to continuously check and reexamine these stages to ensure good quality training data\cite{b12}.\\
\midrule
5 & Data validation $\longrightarrow$  Model Training & \scriptsize	Since most machine learning models are complex black boxes, validating them involves constantly assessing and monitoring data to pinpoint issues and evaluate their quality. Making it problematic to maintain because the model is continuously updated using new data in an online learning system\cite{b12}. Without proper data validation strategies, errors or undesired behaviour in data could cause degradation in training performance and even training-serving skew , which in turn may lead to this Repetitive Maintenance anti-pattern. \\
\midrule
6 & HPO $\longrightarrow$  Model Training & \scriptsize	Configuring the right sets of Hyperparameter is a prolonged and manual process; it is often done trial and error without expert knowledge. In general, Models' performance, efficiency and rate of convergence of models are all dependent on HPO\cite{b22,b23}. Wrong choice in these parameters often directly influences the learnability and rate of coverage in the Model training stage, which may often lead to retraining the model with different parameters, which is a waste of resources and time.\\
\midrule
7 & Model Training $\longrightarrow$ Model Deployment & \scriptsize	 The choice of modelling and training technique affects how a model performs and where it is deployed. For example, incremental learning is more accurate for highly fluctuated and adapted systems, while retrained is better for stable systems\cite{b18}.Challenges in deployment also arise when transitioning from prototype to production, where glue code, ad hoc and brittle pipelines have to be managed and require setting up monitoring and logging capabilities\cite{b9,b10,b27}.\\
\midrule
8 &  Model Deployment $\longrightarrow$ Model Testing & \scriptsize Trained models need to be deployed and integrated with other models or applications. However, due to different OS and hardware environments, constraints like power, memory and vendor-specific optimisation packages and libraries\cite{b26,b28} make it more demanding when developing integration tests and handling edge cases for different environments adding to the complexity of maintaining ML testing.\\
\midrule
9 & Data Validation $\longrightarrow$ Model Drift & \scriptsize	Changing data occurs when fluctuations in data collection are unavoidable or due to data seasonality. It is not feasible to create a data validation algorithm that can detect all types of these drifts \cite{b17,b19}. Researchers have many methods to deal with drifts, but none is perfect. Most of these methods are expensive to implement because they require knowledge of drift detection algorithms, engineering the solution into existing pipelines and the ongoing  maintenance required in detecting new drifts\cite{b17}.\\
\midrule
10 & Model Drift $\longrightarrow$  Model Testing & \scriptsize	ML systems are stochastic and data-dependent, making them susceptible to Data and Concept Drift which leads to rapid obsolescence of input and expected output parts of test cases and creates a moving target, and have fundamentally different nature and construction compared to traditional software, posing new challenges for detecting model drifts and authoring/maintaining unit test and regression tests resulting in the anti-pattern of Repetitive Maintenance\cite{b33,b34,b35,b37}.\\
\midrule
11 & Model Testing $\longrightarrow$ Model Monitoring & \scriptsize	ML systems influence their behaviour over time and may lead to hidden feedback loops where the input to the model is being indirectly adjusted to influence its behaviour. Testing and monitoring for hidden feedback loops, data errors, performance metrics and drifts in an evolving data-dependent system is a challenging problem and requires understanding on what are the data and model quality attributes to monitor and how to alert them, making it susceptible to Repetitive Maintenance anti-pattern \cite{b19,b24}.\\
\midrule
12 & Model Testing $\longrightarrow$ Model Governance & \scriptsize In a high-risk ML application, many steps like defining quality metrics and requirements specifications are cross-disciplinary efforts and require rigorous formal verification tests and also testing model quality in real-life setting \cite{b19} to ensure regulatory and ethical compliance. However, ML testing is faced with many challenges because of its stochastic nature and data dependency, thereby impacting the Model Governance when running these suites of verification tests in real-life settings, consequently leading to Repetitive Maintenance.\\
\midrule
13 & Model Governance $\longrightarrow$ Model Monitoring & \scriptsize Model owners are usually responsible for the documentation and risk management of their models, as well as ensuring their compliance with all regulations and identifying the metrics which have to be monitored\cite{b26}. Engineers have to build custom solutions in order to monitor the ML application effectively and provide visual tools and implement access privileges for team members \cite{b19,b26,b27} which is a huge undertaking and requires constant maintenance to ensure good quality model governance.\\
\bottomrule
\end{tabularx}
\end{table*}
\section{Implications}
\subsection{Implication for Developer of the ML tools}
From our SLR, it emerged that many ML workflows demand provenance tracking, publishing of ML models and their artefacts, tracking data transformations, querying and storing intermediate steps \cite{b4,b30,b31}.There is a lack of standard tools and methods, which allows ML tools developers to build solutions based on these technology gaps.\\
Many ML projects fail at the prototyping stage because setting up infrastructure for deployment and maintenance requires integration and management of glue code, ad-hoc pipelines, and data monitoring. So there is a need for additional tooling and frameworks to facilitate the transition from prototype to production environments where the model can easily be maintained and updated \cite{b9,b27,b38}.\\ There is also a lack of ease to use language-independent tools and solutions that can be integrated with any existing frameworks \cite{b17, b24,b41,b45}. As a result, we see a chance for developers to develop tools and solutions, in particular improving data preprocessing \cite{b7}, data management \cite{b3}, data validation \cite{b12}, Model drift detection \cite{c17} and Model monitoring \cite{b19,b24}.\\ In collaborative or multi-organisational projects, monitoring processes are complex because different teams have different metrics and requirements, especially in terms of governance and regulations and also a lack of standards to communicate about ML issues and their quality\cite{b4,b51}. Therefore, we need patterns of integration to make this work for collaborative ML projects \cite{b54}.
\subsection{Implication for Researchers}
When developing robust and reliable ML systems, developers and researchers face an increasingly difficult challenge. Due to entanglement and data-dependent behaviour, different data processing steps and approaches affect the model's performance differently. It is unclear even for experienced developers how to select between several data processing steps and how they will affect the model's performance \cite{b1,b5}. \\
As a community, ML testing and monitoring face many challenges and open problems \cite{b19,b24,b36}. The concern is that ML systems constantly adapt to new data, creating a moving target and posing a different set of challenges to maintain unit and regression testing than traditional software projects \cite{b33,b34,b35,b37}. Therefore, it is essential to know what critical metrics of data and model to test and monitor and how to alarm them when monitoring ML applications \cite{b19,b24}. More research in ML testing and monitoring will benefit the entire ML community.

\section{Conclusion}
In this SLR, we have screened more than 13000 papers and analysed 56 in-depth, compiling a comprehensive catalogue of maintainability challenges and their interdependence affecting ML workflow. Our findings will assist practitioners in understanding maintainability challenges and their impact at different stages of the ML workflow. This will help early identification, avoid costly pitfalls, and develop mitigation strategies. Moreover, we provide directions for tool development and further research to improve the maintainability of ML systems.

\renewcommand\refname{Literature Review Papers }
\makeatletter
\renewcommand\@bibitem[1]{\item\if@filesw \immediate\write\@auxout
    {\string\bibcite{#1}{L\the\value{\@listctr}}}\fi\ignorespaces}
\def\@biblabel#1{[L#1]}
\makeatother

\end{document}